\documentclass{article}
\usepackage[final]{corl_2019} % Uncomment for the camera-ready ``final'' version

\usepackage{times}
\usepackage{epsfig}
\usepackage{graphicx}
\usepackage{amsmath}
\usepackage{amssymb}
\usepackage{comment}
\usepackage{subcaption}
\usepackage{multirow}
\newsavebox{\measurebox}
\newcommand{\mb}{\mathbf}

\newcommand{\GG}{\mathcal{G}}
\newcommand{\VV}{\mathcal{V}}
\newcommand{\EE}{\mathcal{E}}

\newcommand{\VEGs}{VEGs }
\newcommand{\VEG}{VEG }

\newcommand{\atte}{\mathrm{att} }

\renewcommand\footnotemark{}

\title{Graph-Structured Visual Imitation}

\author{Maximilian Sieb*, Zhou Xian*, Audrey Huang, Oliver Kroemer, Katerina Fragkiadaki\\
Carnegie Mellon University \\
{\tt\small \{msieb, xianz1, audreyh, okroemer, kfragki2\}@andrew.cmu.edu} \\
\vspace{-30pt}}
\thanks{\hspace{-16pt}* Indicates equal contribution. This work was partly funded by the Sony Corporation. Maximilian Sieb was supported by the German Academic Exchange Service, the Ulderup Foundation and the German Academic Scholarship Foundation during the course of this research.}
\begin{document}
\maketitle
% \thispagestyle{empty}
% \footnote{all authors are members of Carnegie Mellon University (CMU) {\tt\small \{msieb, xianz1, audreyh, okroemer, kfragki2\}@andrew.cmu.edu}}

\begin{abstract}
%Learning robotic  behaviours by watching humans remains elusive due to the difficult visual scene recognition problem that 
We cast visual imitation as a visual correspondence problem. Our robotic agent is rewarded when its actions result in better matching of relative spatial  configurations for  corresponding visual entities detected in its workspace and the teacher's demonstration. We build upon recent advances in Computer Vision, such as human finger keypoint  detectors, object detectors trained on-the-fly  with synthetic  augmentations, and point detectors supervised by viewpoint changes  \cite{florencemanuelli2018dense} and learn multiple visual entity detectors for each demonstration  without  human annotations or  robot interactions. 
We empirically show that the proposed factorized visual representations of entities and their spatial arrangements drive successful imitation of a variety of manipulation skills within minutes, using a single demonstration and without any environment instrumentation. It is robust to background clutter and can effectively generalize across environment variations between demonstrator and imitator, greatly outperforming unstructured non-factorized full-frame CNN encodings of previous works \cite{Sermanet2017}. %, which require a large amount of videos and robot interactions to learn how to handle clutter and environment variations. %, since entities that do not correspond do not participate in the reward function.
%Furthermore, it is robust across  differences in object poses and locations  since our visual entity detectors can be trained to be robust to such variations. %correspondences are found and matched across  different object instances, different object poses and locations. %, which provides robustness to initial 
 %The proposed multi-scale part-based visual representation  is learnt simply by changing camera viewpoints  and via augmenting frames synthetically. 
% In contrast, unstructured non-factorized full-frame CNN encodings of previous works \cite{Sermanet2017} require a large amount of videos and robot interactions to learn how to handle clutter and environment variations.

 \end{abstract}

% Two or three meaningful keywords should be added here
\keywords{Imitation learning, Robotic manipulation, Reinforcement learning}

\section{Introduction}

Humans learn skills by watching other humans \cite{Kuniyoshi}. The ability to learn from observation ---called visual imitation \cite{pathakICLR18zeroshot} or third-person imitation \cite{DBLP:journals/corr/StadieAS17}---
has always been a much-desired goal in artificial intelligence as a means of quickly programming agents in an intuitive manner, as opposed to  hard-coding their behaviors. 
Visual imitation requires  fine-grained understanding of the demonstrator's visual scene and its changes over time. The imitator then will use its own embodiment and dynamics to cause a similar change in its own environment. 
Visual imitation then boils down to  learning a visual similarity  function between the demonstrator's and imitator's environments, whose maximization---via the imitator's actions---would result in correct skill imitation. This similarity function determines  which aspects of the visual observations are relevant to reproducing the demonstrated  skill, i.e., it defines what to imitate and what to ignore \cite{Schaal_TCS_1999}. 

\begin{figure}[ht!]
\centering
  \includegraphics[width=\textwidth]{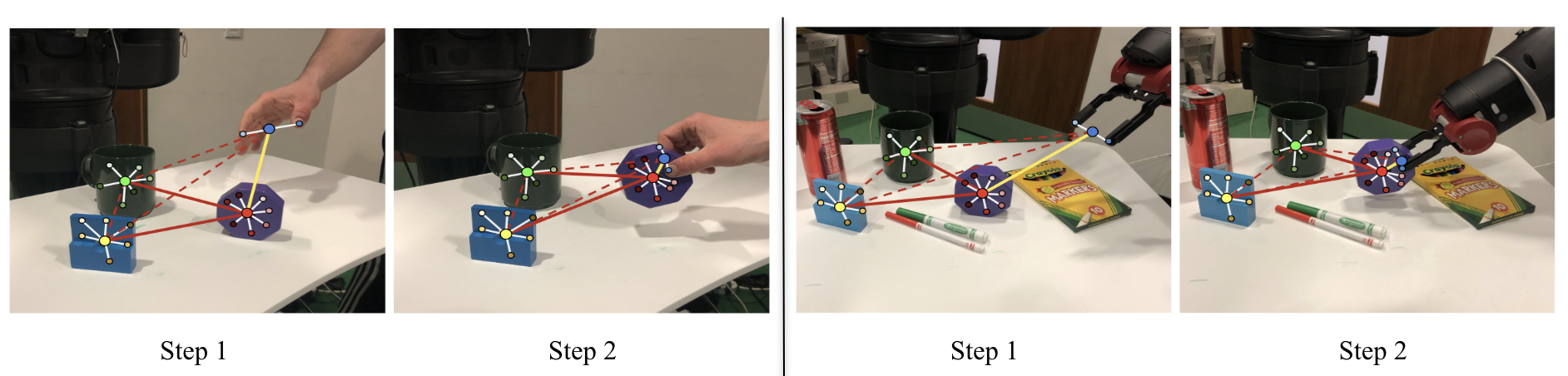}

\caption{ \textbf{Graph-structured visual imitation.} 
We show the VEGs for a human demonstration and robot imitation for two timesteps. Corresponding nodes in the human demonstration (left) and robot imitation (right) share the same color. The graphs  are hierarchical. Edges exist between object,  robot and human hand nodes and point feature nodes, and are added and deleted dynamically over time based on motion saliency, as shown in the figure with solid and dashed lines, respectively.
%Our imitation reward function is determined by the  agreement of the relative spatial arrangements along  corresponding edges present in the two graphs.
Our graph representation is robust to viewpoint variation between demonstrator and executor, and can handle cluttered backgrounds, as illustrated in the figure.
}
\label{fig:intro}
\end{figure}

We propose    hierarchical graph video representations, called Visual Entity Graphs (VEGs), where nodes represent visual entities (objects, parts or points) tracked in space and time,  and edges represent their relative 3D spatial arrangements. Our video graph encoding is based on the  observation that  the appearance  of the scene in some level of abstraction (objects, parts or points) remains constant over time, but the spatial arrangements  of  entities change over time, an observation which follows directly from  the laws of Newtonian physics.  
The proposed hierarchical  visual entity graphs disentangle the ``what" and ``where" of the scene in multiple levels of abstraction:  nodes represent visual entities that  persist over time, and edges within each graph represent their  relative 3D spatial arrangements  that may change over time. For each pair of timesteps, we build two VEGs, one for the demonstrator and one for the imitator. Their nodes are in one-to-one correspondence, as shown in Figure \ref{fig:intro}. 
Our imitation reward function then measures agreement of the relative spatial configurations between corresponding node pairs, and  guides reinforcement learning of manipulation tasks from a single video demonstration  using  a handful of real-world interactions. \\

%We consider a simple motion saliency  heuristic to dynamically add edges between node entities that are in motion in the demonstration video, and focus in this way the reward function to the relevant part of the demonstrator's scene.  
Under the proposed VEG encoding, visual imitation  boils down to learning to detect  corresponding visual entities (objects, parts, or points) between the demonstrator's and imitator's environments. This requires fine-grained visual understanding of both demonstrator's and imitator's  environments.  
The challenge in this visual parsing problem is that objects used by the demonstrator  are often  not included in  the labelled image or object  categories of ImageNet \cite{imagenet_cvpr09} and MS-COCO \cite{DBLP:journals/corr/LinMBHPRDZ14} datasets, making off-the-shelf  pre-trained object-detectors less useful.  
%We achieve scene-conditioned fine-grained visual understanding by marrying three powerful  
We instead opt for scene-specific self-supervised detectors for points and objects.   We use self-supervised  point visual feature detectors trained by viewpoint changes, and  visual detectors of objects and parts trained from synthetically generated images that augment a video at hand. Lastly, we also use human hand keypoint detectors to parse the demonstrator's hand trajectories. The proposed scene-conditioned visual entity detectors establish correspondences between the demonstrator's and imitator's workspaces, despite  differences  in occlusion patterns,  viewpoint changes or  robot-human body visual discrepancies (Figure \ref{fig:intro}). We use the resulting reward function for trajectory optimization \cite{Chebotar2017}, and show that it can imitate a single human demonstration from a handful of real world trials on a Baxter robot.  

In summary, our contributions are as follows: i)
  We propose a what-where hierarchical graph visual encoding for visual correspondence estimation.
    ii) We propose scene-specific point and object visual detectors, as well as human hand detectors. 
   To the best of our knowledge, this is the first work that uses human finger visual detectors as opposed to environment instrumentation \cite{Schmidt2015} to track the demonstrator's hand for visual imitation learning of object manipulations.
    iii) We imitate using a single demonstration, without any robot random exploration as in \cite{DBLP:journals/corr/NairCAIAML17,pathakICLR18zeroshot}, or any data of the robot performing the task as in \cite{Sermanet2017}. We do so without ever having access to expert actions. 
    iv) We show imitation results on a real robotic platform.

We compare our proposed representation against full-frame image encodings of previous works \cite{Sermanet2017,DBLP:journals/corr/NairCAIAML17} that do not use a what-where  decomposition during matching. Our experiments suggest they require a very large number of video examples of humans and robots executing the task to acquire  generalization abilities similar to our method.   
%By examples, we mean videos of humans and robots executing the task, which is data hard to collect. 
They fail to imitate the demonstrated skill most of the times, as we show in our experiments. When humans imitate fellow humans, they are equipped with excellent  visual detectors, visual feature extractors, and motion estimators as opposed to learning those from scratch for every new task. 
We opt for a similar transfer of machine vision knowledge during imitation for robotic agents. 

% Our code and accompanying videos are included in the supplementary materials.
Our code and videos are available at \href{https://msieb1.github.io/visual-entity-graphs}{https://msieb1.github.io/visual-entity-graphs}.
\begin{comment}

Direct matching of pixel intensities is not a meaningful measure of  similarity \cite{DBLP:journals/corr/JohnsonAL16} as it is easily spoiled by difference in viewpoints, human body versus robot body parts, illumination changes, changes in object poses, etc.. 
Numerous previous works manually 
Few recent works attempt instead to   \textit{learn}  such visual similarity by training  image (or short image sequence) feature  embedding in which L2 distances capture the desired notion of similarity , use unsupervised time-contrastive losses where  %  that captures the desired notion of similarity, %as opposed to manually design it, using convolutional networks  and self-supervised objectives. 
%For example, Nair et al.  \cite{DBLP:journals/corr/NairCAIAML17} train  inverse and forward visual models, Sermanet et al. \cite{Sermanet2017}  employ   multi-view time-contrastive losses to train image embedding functions so that near-in-time  image frames have small embedding distance,  
far-in-time image frames have high embedding distances, images across different views of the same time instance also have small embedding distance. 
%``pull together" image encodings of nearby frames across different views and ``push apart" image encodings of far  in time frames. 
\end{comment}

\section{Related  Work} \label{sec:related}

\paragraph{Visual imitation learning.} Imitation learning  addresses  
the problem of learning skills by observing expert demonstrations
%, and is considered  an important topic in the field of robotics due to its potential to highly expedite trial-and-error learning  
\cite{Schaal_TCS_1999}.  However, 
most previous approaches assume that expert demonstrations are given in the workspace of the agent, (e.g., through kinesthetic teaching or teleoperation  \cite{Hussein:2017:ILS:3071073.3054912,Argall:2009:SRL:1523530.1524008}) and the actions/decisions of the expert can be imitated directly  \cite{DBLP:journals/corr/abs-1710-04615}. %, either with behavioral cloning (BC) \cite{DBLP:journals/corr/abs-1710-04615}, inverse reinforcement learning (IRL) \cite{Ng:2000:AIR:645529.657801}, or generative adversarial imitation learning (GAIL) \cite{NIPS2016_6391}.  
%\paragraph{Visual imitation}
Imitating humans based on visual information is  much more challenging  due to the difficult visual inference needed for fine-grained activity understanding \cite{DBLP:journals/corr/StadieAS17}. In this case, a mapping between observations in the demonstrator space to observations in the imitator space is required and is essential for successful imitation  \cite{Nehaniv:2002:CP:762896.762899}. 
% but, most importantly, inference of the ``reward", that is, goal of the behaviour, that the agent will attempt to match by self-practise.
Numerous works bypass the difficult perception problem by using special instrumentation of the environment, such as AR tags, to read off object and hand 3D locations and poses during video demonstrations, and use rewards based on known 3D object goal configurations \cite{DBLP:journals/corr/abs-1709-10087,DBLP:journals/corr/KroemerS16}.  %the need for special instrumentation  hinders scaling-up learning from demonstration.  
Other works  
%does not attempt to learn rewards for actions \cite{DBLP:journals/corr/RahmatizadehAB16},  
use hand-designed reward detectors that work only in restrictive scenarios \cite{MullingKKP2012_2}. Direct matching of pixel intensities is not a meaningful measure of  similarity \cite{DBLP:journals/corr/JohnsonAL16} as it is easily spoiled by difference in viewpoints, human body versus robot body parts, illumination changes, or changes in object poses.

%\paragraph{Perceptual reward learning}
Recent approaches to attempt instead to   \textit{learn}  such visual similarity by training and  matching  whole image feature embeddings directly, and avoid  explicit extraction of the scene structure in terms of objects and their 3D poses. 
%or learn full image or image sequence convolutional image embeddings using  time-contrastive losses \cite{Sermanet2017,dwibedi2018learning}, frame prediction \cite{}, or other self (un)-supervised losses \cite{}. 
%However, the data used to train such image embeddings are both human video demonstrations as well as  robot executions of the task, or parts of the task, so that the neural network embedding  function learns to be robust to the presence of the robotic gripper or human hand. Yet, the requirement of the robot executing the task beats the purpose of visual imitation, and brings it closer to kinesthetic teaching. The same holds for recent work of \cite{DBLP:journals/corr/abs-1806-09655}, which learns an image encoding via frame prediction using robot's execution data, albeit the paper title.  By construction, our proposed  graph representations can effectively ignore background clutter or irrelevant parts of the scene \cite{} by not including the corresponding nodes in the VEGs. Moreover, our visual entity detectors can effectively generalize across changes of object locations and poses, as we empirically demonstrate. 
Numerous objectives have been proposed to learn full image or image sequence convolutional image embeddings, such as   multiview invariant and time-contrastive objectives in \cite{Sermanet2017,dwibedi2018learning}, forward and inverse dynamics model learning in \cite{pathakICLR18zeroshot,DBLP:journals/corr/AgrawalNAML16}, or reconstruction and temporal prediction objectives in \cite{DBLP:journals/corr/FinnTDDLA15,DBLP:journals/corr/WatterSBR15}. Work of  \cite{DBLP:journals/corr/abs-1802-04181} provides an overview of common objectives and inductive biases for state representation learning. However, the data used to train such image embeddings are both human video demonstrations as well as  \textit{robot executions of the task}, or parts of the task, so that the neural network embedding  function learns to be robust to the presence of the robotic gripper or human hand. Yet, the requirement of the robot executing the task beats the purpose of visual imitation, and brings it closer to kinesthetic teaching. The same holds for recent work of \cite{DBLP:journals/corr/abs-1806-09655}, which learns an image encoding via frame prediction using robot's execution data, albeit the paper title. Instead, learning our graph video encoding does not require robot executions.

%A single RGB frame or an entire sequence of frames is encoded into an embedding vector using a combination of forward and inverse dynamics model learning in \cite{pathakICLR18zeroshot,DBLP:journals/corr/AgrawalNAML16},  multiview invariant and time-contrastive metric learning in \cite{Sermanet2017}, or reconstruction and temporal prediction objectives in \cite{DBLP:journals/corr/FinnTDDLA15,DBLP:journals/corr/WatterSBR15}.  Work of  \cite{DBLP:journals/corr/abs-1802-04181} provides an overview of common loss metrics and inductive biases for state representation learning. 

 %Having learnt a latent state feature space, imitation is carried out by iterative application of the inverse model in \cite{pathakICLR18zeroshot,DBLP:journals/corr/AgrawalNAML16}, or via trajectory optimization in  \cite{Sermanet2017}. 
%, where the state is the combination of the latent visual state embedding vector and the robot configuration.
%Similar to \cite{Sermanet2017}, we use a  trajectory optimization method \cite{Chebotar2017, theodorouAISTATS2010} to imitate a human visual demonstration. However, instead of learning  a convolutional neural network to embed an image into a single 1D feature  vector, we learn to detect salient objects and parts, and establish the matching of the relative distances of corresponding visual entities in 3D space in demonstrator's and learner's environments as a learning signal. % spatial arrangements. 

Similar to our work, work of \cite{Sieb2019a} also uses 3D spatial object arrangements to guide visual imitation of manipulation tasks. However, they do not consider human keypoints or any entities finer than objects, which suggests their method can only imitate simple translation tasks, where object pose is not relevant (e.g., they cannot handle rotation). %We additionally use human finger keypoint detectors and point-based image features to handle grasping and other fine-grained object manipulation tasks.
Work of \cite{TOGSFV} utilizes human pose detectors to imitate 3D human motion extracted from YouTube videos of acrobatic activities in simulation. However,  no contact with objects is considered, and  imitation of human motion only happens in a simulated agent.   
%, but demonstrations are limited to physically simulated humanoids and robots. 
In comparison, we do not imitate motion alone, but rather, we  carry out a desired manipulation of the environment. Human hands are part of the graph we attempt to create, but so are the surrounding objects in the scene.

\vspace{-8pt}
 \paragraph{Scene graphs, object-centric reinforcement learning, and relational neural networks.} Representing a visual  observation in terms %nodes 
 %Utilizing graph encodings of a visual scene in terms 
 of objects or 
 parts and their pairwise relations has been found beneficial for generalization of action-conditioned scene dynamics  \cite{DBLP:journals/corr/KanskySMELLDSPG17,DBLP:journals/corr/FragkiadakiALM15, DBLPjournals/corr/BattagliaPLRK16},   body motion and person trajectory forecasting \cite{DBLP:journals/corr/JainZSS15,7780479}, and reinforcement learning \cite{Diuk:2008:ORE:1390156.1390187}. Such graph-encodings have also been used to learn a model of the agent \cite{graphnet}, and use it for model-predictive control in non-visual domains. Work of Devin et al. \cite{Devin2017} uses pretrained object detectors and learns attention over the obtained detection boxes, which are incorporated as part of the state representation for policy learning. The graph representation we propose in this work not only employs explicit attention  to relevant objects, parts, and points, but also preserves their correspondence in time, i.e., the detectors \textit{bind} with specific objects, parts, points over time.

\section{Formulation}

We encode a  demonstration video of length $T$  provided by the human expert and an imitation video of the same length $T$ provided by the imitator in terms of two graph sequences
$\GG_D^t=\{\VV^t_{D},\EE^t_{D}\}, t=1 \cdots T$ and $\GG_I^t=\{\VV^t_{I}, \EE^t_{I}\}, t=1 \cdots T$, respectively. We omit the subscript $D$ or $I$ when either it is clear from the context or it is not important to which workspace we are referring to. %, for clarity.  
%$\GG_D^t=\{\VV^t_{D, o},\EE^t_{D, o}, \VV^t_{D, p},\EE^t_{D, p}, \VV^t_{D, h},\EE^t_{D, h}\}, t=1 \cdots T$ and $\GG_I^t=\{\VV^t_{E, o}, \EE^t_{E, o},\VV^t_{E, p}, \EE^t_{E, p},  \VV^t_{E, h}, \EE^t_{E, h}\}, t=1 \cdots T$, 
A node $\VV^t_i$ corresponds to the $i$th visual entity and its respective 3D $(X,Y,Z)$ world coordinate $\mathbf{x}^t_i$, and an edge $\EE^{t}_{(i,j)}$ correspond to  a 3D spatial relation  between two node entities to be preserved during imitation.  We define a visual entity node $\VV^{t}_i$  to be any object, object part, or point that can be reliably detected  in the demonstrator's \textbf{and} imitator's workspace. All nodes are in one-to-one correspondence between the demonstrator and imitator graph, as shown in Figure \ref{fig:intro}.   An entity can dynamically appear and disappear over time. We only require each entity associated with the demonstration sequence to have a corresponding entity in the imitation sequence. 

We consider three types of nodes: object nodes $\VV_o$, point nodes $\VV_p$, and hand/robot nodes $\VV_h$.
%correspond to \textit{object}, \textit{point}, and \textit{hand/end-effector} node entities, respectively.
An object node represent any rigid or non-rigid object that constitutes a separate physical entity in the world, while a point node represents any 3D  physical point on an object, as seen in Figure \ref{fig:intro}. Hand and robot nodes represent the human wrist 3D location, and the robotic gripper center 3D location, respectively.
%manipulation agent in the scene.
We do not consider edges between point nodes. Rather,  each point node is connected only to the object node it is part of. In that sense, our graphs are hierarchical. 
 
Our cost function at each time step $t$ measures visual dissimilarity across the demonstrator and imitator graphs $\GG^t_D$ and $\GG^t_I$ in terms of relative spatial arrangements of corresponding entity pairs, as follows:

\begin{equation}
\label{eq:reward}
    \mathcal{C}(\GG^t_D,\GG^t_I)= \sum_{i,j, i <j}  w(\EE^t_{(i,j)}) \cdot  \atte(\EE^t_{(i,j)}) \cdot \Bigl\| \left( \mathbf{x}_{D,i}^{t}-\mathbf{x}_{D,j}^{t}\right)     - \left( \mathbf{x}_{I,i}^{t} - \mathbf{x}_{I,j}^{t}\right)\Bigr\|, 
\end{equation}
\begin{figure*}[!t]
\centering
  \includegraphics[width=1.0\textwidth]{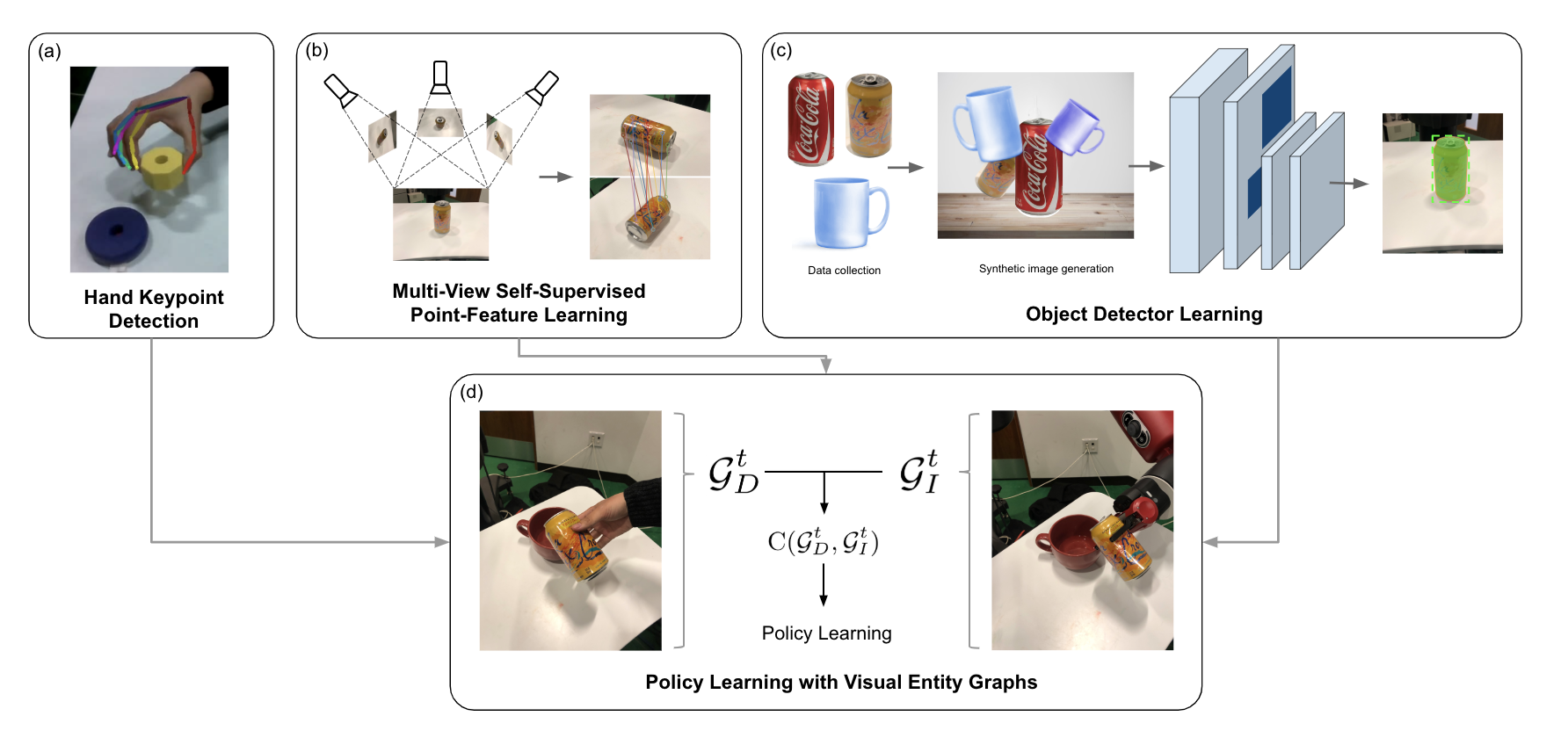}
\caption{\textbf{Detecting visual entities.}
%\todo{new great figure to match the caption}
We use human hand keypoint detectors, multi-viewpoint feature learning, and synthetic image generation for on-the-fly object detector training from only a few object mask examples. Using a manually designed mapping between the human hand and the robot, the visual entity detectors can effectively bridge the visual gap between demonstrator and imitator environment, are robust to background clutter, and generalize across different object instances. 
}
\label{fig:entities}
\end{figure*}
where $\atte(\EE^{t}_{(i,j)}) \in \{0,1\}$ is a binary   attention function that determines whether a particular edge is present depending on the motion of the corresponding nodes, and  $w(\EE^{t}_{(i,j)}) \in \mathbb{R}$ denotes edge weights. We tie weights across all edges of the same type, namely, object-hand edges, object-object edges and object-point edges. They are hyper-parameters of our framework and we set them them empirically. Learning to adjust those weights per task is an interesting and straightforward direction which wee leave for future work.
 %The cost function encodes the relative spatial arrangements in 3D between all entities for which the binary attention function $\atte{(\EE^{t}_{(i,j)})}$ is 1. % for the associated edge $\EE^{t}_{(i,j)}$. 

 %, except for the  pushing through a 90 degree angled trajectory, in which we increased. %In the pushing task
%We use different weights for object-point edges and object-object edges. %, as detailed 
%$pa(\cdot)$ in the superscript maps a point node to its parent object node, and $\alpha$ and $\beta$ are weights to balance the cost, which in our case are simply used to average the cost over all corresponding nodes. The coordinate feature of the \text{hand/end-effector} entity is also weighted by $\lambda$, as mentioned earlier.

\vspace{-10pt}
\subsection{Detecting Visual Entities} 

\vspace{-8pt}
We define a visual entity to be any object, object part, or point that can be reliably detected in the demonstrator's and imitator's workspace.  
 For imitating fine-grained manipulation of an object, inferring the translation of its bounding  box is not enough, rather, the object's 3D pose and deformation needs to be inferred and imitated. 
 A central  design choice in our work is using point feature detectors and motion of the detected points to infer the object's change of pose between demonstrator's and imitator's environments, without additional learning, as opposed to training object appearance features extracted within the object's bounding box to encode object change pose or deformation. We opt for what-where decomposition of the object's appearance, as opposed to whole object box  embedding learning.

 We assume no access to  human annotations that would mark relevant corresponding entities across demonstrator's and imitator's environments. 
We instead train scene-specific object and point detectors for  entities that can be reliably recognized across  demonstrator's and imitator's workspaces, and  human hand keypoint detectors for tracking the human hand. The point detector re-samples points at each step randomly on detected area of objects in demonstration and computes corresponding points in the imitator's view, and is thus robust to partial occlusions. In case of full occlusion, our hand and object detectors use last known location in the past. Thus, our detection pipeline possesses certain robustness to object occlusions and possible detector failures.

\vspace{-8pt}
\paragraph{Human hand keypoint detectors.} We make use of state-of-the-art hand detectors of Simon et al. \cite{simon2017hand} to detect human finger joints, and obtain their 3D locations using a D435 Intel RealSense RGB-D camera.   We rely on forward kinematics and a calibrated camera with respect to the robot’s coordinate frame to detect the  3D locations of the tips of the robot’s end-effector. We map the finger tips of a Baxter robot's parallel-jaw gripper to the demonstrator's thumb and index finger tips.  We detect grasp and release actions by thresholding the distance between the two finger tips of the human during the demonstration of the task. End-to-end approaches such as \cite{Li} rely on large amounts of data to train hand-to-robot correspondences, and are therefore prohibitive in few-shot learning scenarios, whereas our method works thanks to the thousands of labelled hand examples the human hand detectors of \cite{simon2017hand} has seen.

\vspace{-10pt}
\paragraph{Point feature detectors from cross-view correspondence.} An agent that has access to its egomotion and observes a static scene from multiple views can infer visual correspondences across views through  triangulation \cite{hartley2003multiple}. We use these self-generated visual correspondences to drive visual metric learning of deep feature descriptors that are robust to  changes in the object pose or camera viewpoint. After training, we match  such point features across imitator's and demonstrator's environments to  establish correspondence \cite{florencemanuelli2018dense}.   
 %recover such correspondences in the absence of viewpoint information, namely, to learn deep feature descriptors of points that are robust to  changes in the object pose or camera viewpoint. %In our case, such points are pixels in 2D images taken by the cameras.
%\todo{Xian to write details here and an equation, notes: full frame multiview in TCN is too coarse to recover fine-grained details, only within instance supervision, but it has been shown the descriptors learnt to generalize cross-instance}
%The object detectors provide the spatial location of the objects of interest in the scene while we need correspondences at a lower level to infer local object-level transformation, such as orientation change and deformation. Pixel-level correspondences between the demonstration and robot's executions provide rich information of the object's spatial information. Pixels exist as the finest entities in the scene and are used as nodes at the lowest level of granularity in our graph. We follow the approach presented in \cite{florencemanuelli2018dense} to learn descriptors for pixel entities and compute pixel correspondences. 
We collect multiview image sequences of the workspace of the robotic agent in an automatic fashion: we use an RGB-D camera attached to the robot's end-effector and move the camera while following random trajectories that cover many viewpoints of the scene and at various distances from the objects. We use the robot's forward kinematics model to estimate the camera poses via hand-eye calibration, which, in combination with the known intrinsic parameters and aligned depth images, allows for robust 3D-reconstruction of the scene and provides accurate pixel correspondences across different viewpoints. The complete feature learning setup is illustrated in Figure \ref{fig:entities}(b). During training, we randomly sample image pairs  and generate a number of matching and non-matching pixel pairs. We then minimize pixel-wise contrastive loss \cite{schmidt2017self, choy2016universal}, which forces matching pixels to be close in the learned feature spaces, while maintaining a distance margin for non-matching pixels.  This point feature learning pipeline produces a mapping from an RGB image to dense per-point descriptors. 
Even though supervision comes via within-instance correspondences, generalization across different objects is expected due to the limited capacity of the network model. This enables our VEG representation with powerful generalizability to novel objects unseen in demonstration, as shown in Figure \ref{fig:pouring}.
We use ResNet-34 as our backbone, and learn a 4-dimensional point embedding vector for each pixel in the image.

%the hope is the resulting feature descriptors to match across object instances, due to the limited capacity of the network model. We show that indeed this is the case in Figure \ref{fig:pouring}. 

% with the same resolution, and minimize the distance between matching pixels in the descriptor space, while enforcing at least a certain distance threshold for non-matching pixels.
%\todo{explain pixel feature generalizes over objects}

%Note that such metric learning for obtaining point features is performed given only within-instance correspondences, but within each class, generalization across different objects with different geometries and textures comes freely since learning such embedding are expected due the limited VC dimension of the network model. This enables our VEG representation with powerful generalizabilty to learn skills even with objects unseen in the demonstration.

% While such metric learning for pixel features occurs only over correspondences within the same object, generalization across objects of different geometries and textures in the same class comes freely. \todo{backed up citation?}This is because such embeddings are expected due to the limited VC dimension of the model. Our VEG representation is thus endowed with powerful generalizability that allows it to imitate tasks even with objects unseen in the demonstration. 

\vspace{-10pt}
\paragraph{Synthetic data augmentation.} We use background subtraction to propose object 2D segmentation masks, and train a 
visual  detector  for each mask  using synthetic data augmentation, as shown in Figure \ref{fig:entities}.   %similar, to \cite{Sieb2019a}. %We use the  state-of-the-art deep visual detector architecture of Mask R-CNN \cite{DBLP:journals/corr/HeGDG17}. 
%We create synthetic data by rotating, translating and random-cropping the image, as well as  pasting  random object masks, and recovering the groundtruth (box or mask or point correspondence) with out detector or metric learner. 
Specifically, we create a large synthetic dataset by translating, scaling,  rotating and changing the pixel intensity of the extracted RGB segmentation masks. The  object masks often partially overlap with one another in the synthetic images. These overlaps help the agent learn to detect amodal object boxes under partial occlusions. Since we generate such images, we  automatically know the groundtruth bounding box and mask that correspond to each object in each image.  We then finetune a Mask R-CNN object detector \cite{DBLP:journals/corr/HeGDG17}---initialized from weights learned under the object detection and segmentation task in MS-COCO---to predict boxes and masks for the synthetically generated images.

\vspace{-8pt}
\subsection{Motion Saliency for Dynamic Graph Construction}

\vspace{-8pt}
%A demonstrator's environment may contain  objects that do not  participate in the demonstration, but are rather distractors. Motion saliency is one heuristic humans use to attend to the right part of the demonstrator's environment \cite{}. %It is believed that humans use simple motion heuristics to attend to the right part of a demonstration \cite{} when imitating others. \todo{+++}
We use similar motion saliency heuristics to  decide dynamically over time what edges $(i,j)$ to consider in our imitation cost function of Eq. \ref{eq:reward}, by setting  $\atte{(\EE^{t,(i,j)})}=1$ to denote edge presence. 
We define an anchor object to be any object in motion, and in the case of no moving objects---e.g., when the demonstrator is simply reaching towards an object---we define the anchor object to be the closest-in-the-future moving object.   
We consider edges between the anchor object node and all other \textit{corresponding} object nodes in the scene, as well as the hand/robot node, and all point nodes that belong to the anchor object.
%, i.e. $\atte{(\EE^{t,(i,j)})}=1$, for all edges
% Such anchor-centric graph structure allows flexible camera viewpoint and is robust against shift in absolution spatial configurations of the objects \todo{what does this mean}.\\
This type of motion attention is a well-established principle that drives human attention when imitating other humans. Work of \cite{Lee2009} uses AR markers to estimate such task relevance based on motion attention. We  simply use our object detection network to track objects in time to infer movement between the different objects in the scene.

\vspace{-8pt}
\subsection{Policy Learning with Visual Entity Graphs}

\vspace{-8pt}
Our goal is the robot to imitate the intended object manipulation task from a single human visual demonstration. 
We formulate this as a reinforcement learning problem, where at each time step the cost is given by  Eq. \ref{eq:reward}. 
We use PILQR \cite{Chebotar2017} to minimize the cost function in Eq. \ref{eq:reward}, a state-of-the-art trajectory optimization method that combines a model-based linear quadratic regulator with path integral policy search  to better handle non-linear dynamics. %, which are in general scenarios vanilla LQR-based approaches struggle with. % PILQR is a state-of-the-art reinforcement learning algorithm that combines, in a principled manner, the model-based generalization of linear quadratic regulators together with path integral policy search's flexibility and ability to handle non-linear dynamics. 
%policy search reinforcement learning. The reinforcement learning is used to demonstrate that the learned rewards can be employed to acquire suitable manipulation policies. 
We learn a  time-dependent policy $\pi_t(\mathbf{u}_t|\mathbf{x}_t;\theta)=\mathcal{N}(\mathbf{K}_t\mathbf{x}_t+\mathbf{k}_t, \mathbf{\Sigma}_t)$  
, where the time-dependent control gains are learned by alternating model-based and model-free updates, where the dynamical model $p(\mathbf{x}_t|,\mathbf{x}_{t-1},\mathbf{u}_t)$ of the \textit{a priori} unknown dynamics is learned during training time. 
The actions $\mathbf{u}$ are defined as the changes in the robot end-effector's 3D position and orientation about the vertical axis, giving a 4-dimensional action space. The state representation---over which we learn linear dynamics--- consists of the joint angles, end-effector position, and the graph configuration of the scene, concatenated into one vector. For $N$ objects, this featurization scheme results in a $3+n_{joints}+\sum_{i=1}^{N_{anchors}}(N_{total}-1)*3+\sum_{i=1}^{N_{anchors}} dim(\phi)$ dimensional state space, where $n_{joints}$ denotes the number of joints of the robot and $\phi$ encodes the chosen node feature for a visual entity in the graph. In our case, we perform uniform sampling of node-specific point features at each time step, and we directly incorporate the averaged pairwise distance of all pixels across demonstration and imitation into the state space, yielding $dim(\phi)=1$.
We further use behavioral cloning to infer opening and closing of the robot gripper by thresholding the distance between the human index finger and thumb during the demonstration.

\section{Experiments}
\label{sec:exp}

\begin{figure}
    \centering
    \begin{subfigure}[b]{.48\textwidth}
        \includegraphics[width=\textwidth]{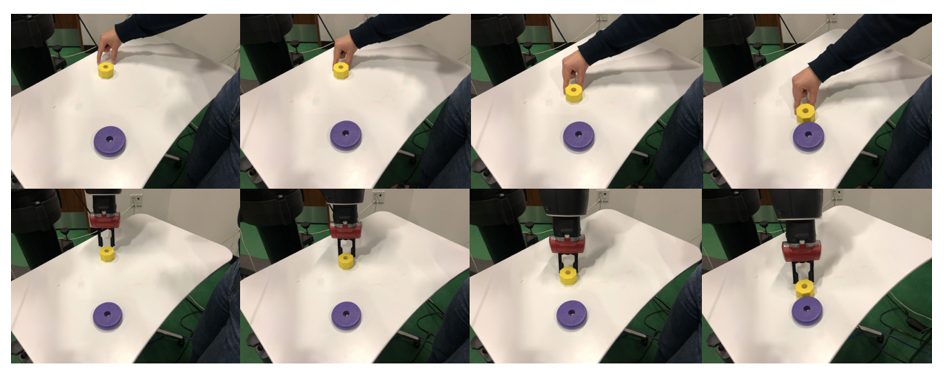}
        \caption{\textit{Pushing}}
        \label{fig:pushing}
        \vspace{4ex}
        \includegraphics[width=\textwidth]{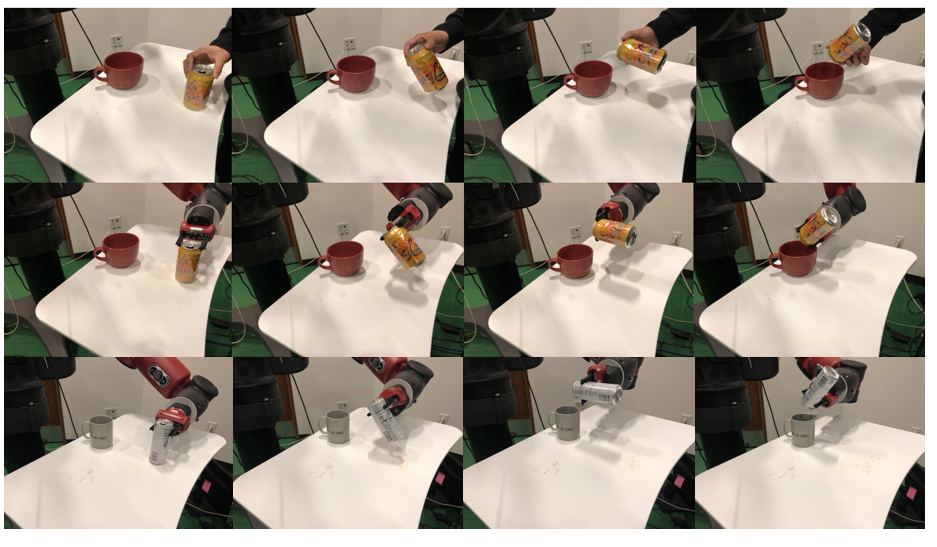}
        \caption{\textit{Pouring}}
        \label{fig:pouring}
    \end{subfigure}
    \begin{subfigure}[b]{.48\textwidth}
        \includegraphics[width=\textwidth]{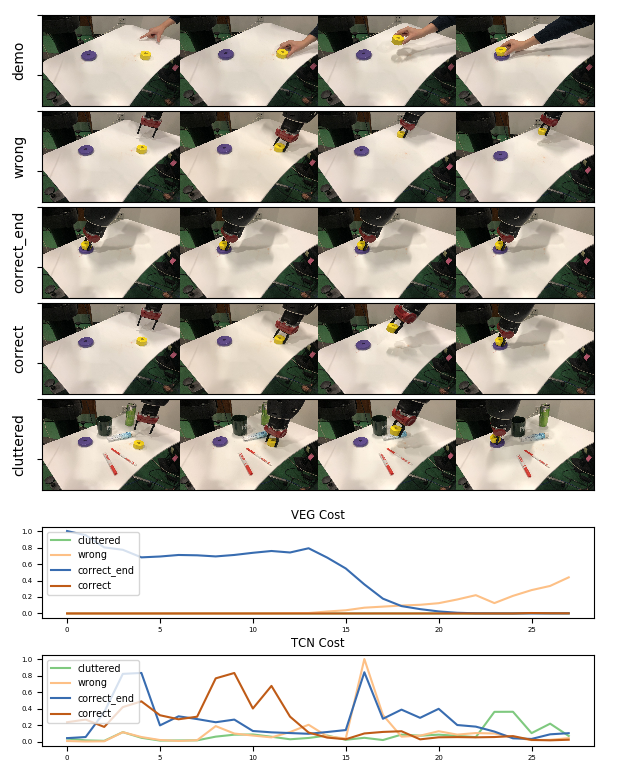}
        \caption{\textit{Stacking} and cost comparison}
        \label{fig:rewardshaping}
    \end{subfigure}\qquad
\caption{Examples of the imitation tasks we consider. (a): \textit{pushing}, (b): \textit{pouring} (c): \textit{stacking} and \textbf{cost comparison of the proposed \VEG cost function  and the TCN embedding cost \cite{Sermanet2017}.} The  \VEG based cost curves are highly discriminative of the imitation quality for the stacking task, in contrast to TCN. 
	Costs are scaled between $[0, 1]$ by dividing by the maximum absolute value of the respective feature space for visualization purposes. \texttt{correct\_end} is simply repeating the last frame of a successful imitation, showing our lost function serves well as an attractor towards the final goal configuration. 
    Note that \texttt{correct} and  \texttt{cluttered} overlap due to both having zero VEG cost, showing that our method is robust to background clutter.}
\label{fig:task}
\end{figure}

We test our visual imitation framework on a Baxter robot. We consider the following tasks to imitate: i) \textbf{Pushing}: The robot needs to push an object while following  specific trajectories. ii) \textbf{Stacking}: The robot needs to pick up an object, put it on top of another one and release the object. iii) \textbf{Pouring}: The robot needs to pour liquid from one container into another.

%Examples for each of the tasks are shown in Figures \ref{fig:task}.
For every task, we train corresponding object detectors using synthetic data augmentation and  point features using multi-view self-supervised feature learning. Note that both processes are fully automated and \textbf{do not require human demonstrations or robot interactions}. Rather, it suffices to use a camera that is setup to move around the scene in a prerecorded fashion

We compare our method   against time-contrastive networks (TCN) of Sermanet et al. \cite{Sermanet2017}. 
We are not aware of other instrumentation-free methods that have attempted single shot imitation of manipulation skills, without assuming extensive pre-training with random actions for model learning \cite{DBLP:journals/corr/NairCAIAML17}.   
%\paragraph{Baseline: time-contrastive image embedding learning}
TCN trains an image embedding network using a  triplet ranking loss  \cite{Hoffer2015}, ensuring that temporally near pairs of frames are closer to one another in embedding space than any temporally distant pairs of frames. %so that RGB frames near in time have small embedding feature distances, and RGB frames far in time have large embedding feature distance. 
In this way, the feature extractor focuses on the \textit{dynamic} part of the demonstration video. 
%Specifically, they use a  triplet ranking loss as introduced in \cite{Hoffer2015}, ensuring that temporally near frames are closer to each other in embedding space than any temporally distant pair of frames. 
We implemented the TCN baseline  using the same architecture as in \cite{Sermanet2017}, which consists of an Inception-v3 model \cite{Szegedy2015}, pretrained on ImageNet, up to the “Mixed 5d” layer which is then followed by two convolutional layers, a spatial softmax layer and a fully-connected layer. The network finally outputs a 32-dimensional embedding vector for the input image. For each imitation task, we train a corresponding TCN using 10 video sequences, 5 human demonstrations and 5 robot executions while performing tasks with relevant objects and environment configuration, together with the human demonstration we provide for learning the policy with VEG. 
With respect to the policy learning with TCN, we use a cost function of the form 
$\alpha||\mb{z}_t-\mb{w}_t||_2^2+\beta\sqrt{\gamma+||\mb{z}_t-\mb{w}_t||_2^2}$, where we choose $\gamma=10^{-5}$, $\alpha=1.0$, and $\beta=0.001$. Here, $\mb{z}_t$ and $\mb{w}_t$ denote the state embedding at each time step for the imitation and demonstration, respectively.
%Our method does not require additional data as TCN does.  
While our method uses only a single human demonstration, 
it does require training  object detection and point-feature networks, although both the data synthesizing for the object detector and the data collection for the point-feature network are fully automated.
To ensure a fair comparison, we put the same amount of time effort into data collection for training the TCN baseline.

Our experiments aim to compare the proposed graph structured encoding against convolutional image encodings of previous work \cite{Sermanet2017} for imitating skills, evaluate the robustness of our method against detectors' failures and occlusions, evaluate its robustness to variability in the objects' spatial configurations \& background clutter and its generalizability  across objects with different  shapes and textures.

\vspace{-8pt}
\paragraph{Reward Shaping.} In Figure \ref{fig:rewardshaping}, we show  reward curves for our method and the TCN baseline for each robot execution, measuring how well the robot is imitating the human demonstration. %,  computed by the proposed \VEG based cost function and the baseline TCN \cite{Sermanet2017}  cost function.
%we consider a human demonstration of a stacking task and  various robot executions under different conditions. We then show a comparison of the corresponding cost curves for each robot execution with the human demonstration,  computed by the proposed \VEG based cost function and the baseline TCN \cite{Sermanet2017}  cost function. 
The horizontal axis denotes time and the  vertical axis denotes imitation cost. The proposed graph-based cost function correctly identifies all correct robot imitations, despite heavy background clutter in the 5th row, and correctly signals the wrong  imitation segments in 2nd and 3rd rows. In contrast, the baseline TCN cost curves are non-discriminative. 
%, and \textit{2)} discriminative enough for effective skill learning
%We compare the cost computed based on VEG and TCN on the \textit{stacking} task, while we provide a single demonstration and evaluate the resulted cost at each timestep for robot's execution under different conditions. The results are visualized in Figure \ref{fig:rewardshaping}. 
%We show in the top row the successful demonstration, followed by successful imitation carried out by the robot and their corresponding cost shaping curve. Note our VEG encoding, compared to TCN, provides much more informative cost signal for the correct imitation, and more robust under different variations, which 
Highly discriminative cost curves are critical for effective policy learning, which we discuss right below.

\vspace{-8pt}
\paragraph{Task - Pushing.} In this task, the robot needs to push the yellow octagon towards the purple ring following the trajectory showed by the demonstrator (Figure \ref{fig:pushing}). We evaluate three task variations: i) \textit{straight-line:} 
    Pushing the object following a straight line. ii)  \textit{straight-line-grasped:}
    Moving the yellow octagon along a straight line while it is being grasped.
iii) \textit{direction-change:}
    Pushing the yellow octagon along a trajectory with a sharp direction change of 90 degrees. Imitating such direction change requires the robot to change the point of contact with the object during pushing. For straight-line pushing, we attempted imitation in two different environments,
%run PILQR for both 
one with a yellow octagon block, and one with a smaller orange square block  to evaluate the generalization capability of our graph-based framework across  objects with different geometries. Essentially, we substitute the object detector for the small orange square block in the second case for the detector of the yellow octagon block. 

\begin{table}[ht!]
\centering

  \includegraphics[width=\linewidth]{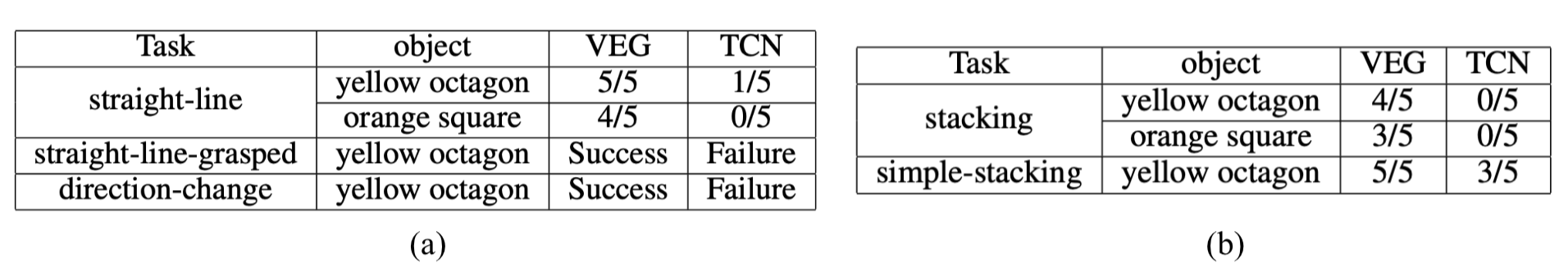}
  \caption{Success rates for the pushing and stacking tasks.}
  \label{tab:comp}
\end{table}

% \begin{table}[ht!]
% \centering
% \begin{tabular}{|c|c|c|c|}
% \hline
% Task & object & \VEG & TCN \\ \hline
% \multirow{2}{*}{straight-line} & yellow octagon & 5/5 & 1/5 \\ \cline{2-4} 
%  & orange square & 4/5 & 0/5 \\ \hline
% % \multirow{2}{*}{straight-line-grasping} & yellow octagon & Success & Failure \\ \cline{2-4} 
% %  & orange square & Success & Failure \\ \hline
% straight-line-grasped & yellow octagon & Success & Failure \\ \hline
% % \multirow{2}{*}{direction-change} & yellow octagon & Success & Failure \\ \cline{2-4} 
% %  & orange square & Success & Failure \\ \hline
% direction-change & yellow octagon & Success & Failure \\  \hline
% %  & orange square & Success & Failure \\ \hline
% \end{tabular}
% \caption{Success rates for the pushing tasks under different settings.}
% \label{tab:push}
% \end{table}

\vspace{-5pt}
In order to test  the robustness of \VEG to variations in the objects' spatial configurations, for the basic straight-line pushing task, we randomly perturb the starting configurations of all the objects and the robot's end-effector within a norm ball of diameter 6cm, and run the policy training 5 times for each object. We consider the task solved if the robot is able to push the object  within 1cm of the desired target position. We report the success rate for each case in Table \ref{tab:comp}(a). The robot successfully solves the task for all 5 runs for the original yellow octagon, and for 4 runs out of 5 for the smaller orange square, demonstrating robustness against perturbation in objects' spatial configuration and strong generalization over objects with novel geometries. TCN failed in almost all runs, and in case for simple straight-line pushing, it failed to push the object into the target region even with significantly more iterations. 
For the straight-line-grasped task, the robot is forced to keep the gripper closed. Hence, the robot cannot simply hard-imitate the hand trajectory, but rather needs to follow a trajectory that differs from the human demonstrator to successfully solve the task. In this sense, the hand trajectory of the human serves as a weak guiding signal during the policy learning. 
The robot solves both straight-line-grasped and direction-change pushing after 8 iterations of trajectory optimization. Note that we increased the hand edge weight $w$ to account for stronger guidance of the hand during the direction-change task. TCN fails to solve any of the latter two tasks, generally performing  poorly in tasks where the robot does not have continuous contact with the object.

\vspace{-8pt}
\paragraph{Task - Stacking.}
In this task, the robot needs to reach and grasp the yellow octagon and position it on top of the purple ring, as shown in the 1st and 4th rows in Figure \ref{fig:rewardshaping}. As in our pushing experiments, we randomize the starting configurations of the scene and the robot, and evaluate our method on this augmented setup for stacking both yellow octagon and the orange square. We report the results in Table \ref{tab:comp}(b). \VEGs enables robust policy learning for the yellow octagon, and shows generalization towards objects with novel geometries that have not been observed during demonstration. Being flatter than the octagon, the orange square is much harder to  grasp, yet the robot is able to learn the skill successfully most of the times. 
The TCN baseline %performs poorly because its use of image-level embedding 
is not able to learn a successful grasping action, which requires precise and coordinated end-effector movement. We then evaluate TCN on a simpler task: stacking an already-grasped yellow octagon (simple-stacking in Table \ref{tab:comp}(b)). The TCN performs reasonably well on this task, demonstrating its ability to imitate smooth trajectories, but having difficulty to imitate  fine-grained grasping actions. 
% \begin{table}[ht!]
% \centering
% \begin{tabular}{|c|c|c|c|}
% \hline
% Task & object & \VEG & TCN \\ \hline
% \multirow{2}{*}{stacking} & yellow octagon & 4/5 & 0/5 \\ \cline{2-4} 
%  & orange square & 3/5 & 0/5 \\ \hline
% simple-stacking & yellow octagon & 5/5 & 3/5 \\ \hline
% \end{tabular}
% % \caption{Success rate for the stacking task for our method and baseline with different starting configurations.}
% \label{tab:stack}
% \end{table}

\vspace{-8pt}
\paragraph{Task - Pouring.}
In this task, the robot needs to simultaneously translate and rotate the can to reach the desired orientation above the mug, as shown in 2nd row in Figure \ref{fig:pouring}.  Using the same pouring demonstration, we additionally evaluate generalizability of \VEGs by using a novel object with a different shape and  texture during imitation (Figure \ref{fig:pouring} 3rd row). 
Trajectory optimization converges after 10 iterations, and the robot solves the task for all objects. The TCN baseline is able to move the can along the trajectory, but fails to rotate it to the right configuration for successful pouring.

\vspace{-8pt}
\paragraph{Discussion/Limitations.} Imitating the human hand critically improves performance during  reaching and grasping motions in human demonstrations, which cannot be easily inferred from any object motion. While our method can deal with partial occlusions during imitation, we currently do not attempt tracking of fully occluded objects over a prolonged period of time.
%Rather,  (fully) occluded visual entities are dropped from the graph. This is particularly the case for pixel or object part entities that get occluded much more    often than larger object-level entities. For pixel-level entities, we randomly re-sample 500 pixels in the object mask at each time step. This big number of entities ensures the reward signal is consistent across rollouts and time steps. 
In light of this, a clear avenue for future work is learning to track through occlusions, e.g., by using temporal visual memory from which the reward graph can be extracted, as opposed to relying on visual frames only.  
Active vision can also be used both to undo occlusions during imitation, and to observe the demonstrator from the most convenient viewpoint for the imitator. 
Another limitation is the lack of any prior model information, which would accelerate the policy search. An interesting avenue for future work would be learning a model over the motion of such visual entities, and use it as a prior for better exploration during policy search.

\section{Conclusion}
We proposed encoding video frames in terms of visual entities and their spatial relationships, and we used this encoding to compute a perceptual cost function for  visual imitation. 
Between end-to-end learning and engineered representations, we combine the best of both by incorporating important inductive biases regarding  object fixation and motion saliency in our robot imitator. Experimental results on a real robotic platform demonstrate the generality and flexibility of our approach.  Quoting the authors of \cite{Battaglia2018RelationalIB}, ``just as biology uses nature and nurture cooperatively, we reject the false choice between ``hand-engineering" and ``end-to-end" learning, and instead advocate for an approach which benefits from their complementary strengths."

{
\setlength{\bibsep}{2pt}
\small
\bibliography{egbib,refs}
}
\newpage
\appendix
\section{Hyperparameters for Experimental Section}
This is a comprehensive list of the hyperparameters used for our experiments. We note that this list might be non-exhaustive in the sense that we do not list hyperparameters which were copied over from their respective source implementation or if they are not detrimental to the performance our method to our knowledge. For example, we refrain from listing parameters such as the number of Anchor Scales used for training the Mask R-CNN architecture since this can be inferred from their source implementation.

% \paragraph{Data Collection:}
% \begin{enumerate}
%     \item Sample rate for D435 cameras: 25 fps
%     \item Image resolution: 640x480
%     \item Clipping value for maximum registered depth: 1.5 m
% \end{enumerate}

% \paragraph{Point Feature Learning Hyperparameters:}

% \paragraph{Mask R-CNN Hyperparameters:}
% \begin{enumerate}
%     \item learning rate: 0.0001
%     \item Steps per epoch: 1000
%     \item Images per batch: 2
% \end{enumerate}

\paragraph{\textit{Pushing} Task:}
\begin{enumerate}
    \item Action space: 4-dimensional (XYZ + rotation)
    \item Time step length: 0.4 sec
    \item Episode length: 30
    \item Cost weight hand edges: 1.0 (50.0 for direction-change variant)
    \item Cost weight object edges: 1.0
    \item Cost weight point edges: 0.0 (not used)
    \item Number of rollouts per update cycle: 8
    \item Controller gains: XYZ: 0.001, rotation: 0.02
\end{enumerate}

\paragraph{\textit{Stacking} Task:}
\begin{enumerate}
    \item Action space: 4-dimensional (XYZ + gripper state)
    \item Time step length: 0.4 sec
    \item Episode length: 30
    \item Cost weight hand edges: 1.0
    \item Cost weight object edges: 1.0
    \item Cost weight point edges: 0.0 (not used)
    \item Number of rollouts per update cycle: 8
    \item Controller gains: XYZ: 0.001
\end{enumerate}

\paragraph{\textit{Pouring} Task:}
\begin{enumerate}
    \item Action space: 4-dimensional (XYZ + rotation)
    \item Time step length: 0.4 sec
    \item Episode length: 20
    \item Cost weight hand edges: 1.0
    \item Cost weight object edges: 1.0
    \item Cost weight point edges: 1.0 
    \item Number of rollouts per update cycle: 10
    \item Controller gains: XYZ: 0.00125, rotation: 0.02
\end{enumerate}

\end{document}